\PassOptionsToPackage{table}{xcolor} % ensure a single, consistent xcolor option
\RequirePackage{amsmath}
\documentclass[runningheads]{llncs}

\usepackage[T1]{fontenc}
\usepackage{microtype}

\usepackage{graphicx}
\usepackage{booktabs}     % (loaded once)
\usepackage{array}
\usepackage{xcolor}
\usepackage{colortbl}

\usepackage{listings}
\usepackage{appendix}
\usepackage{svg}          % note: arXiv cannot run Inkscape; prefer PDF/PNG includes
\usepackage{amssymb}
\usepackage{multirow}
\usepackage{threeparttablex}
\usepackage{enumitem}
\begin{document}
\title{RAGuard: A Novel Approach for in-context Safe Retrieval Augmented Generation for LLMs}
\author{Connor Walker\inst{1,2}\orcidID{0009-0008-8181-1644} \and
Koorosh Aslansefat\inst{1,2}\orcidID{0000-0001-9318-8177} \and
Mohammad Naveed Akram\orcidID{0000-0002-0924-5536} \and
Yiannis Papadopoulos\inst{1,2}\orcidID{0000-0001-7007-5153}}
\authorrunning{C. Walker et al.}
\institute{University of Hull, Cottingham Road, Hull HU6 7RX, UK \and
AURA CDT, Hull UK
\email{C.Walker-2018@hull.ac.uk}\\
\url{https://www.deis-hull.com/connor-walker}\\}
\maketitle              % typeset the header of the contribution
\begin{abstract}
Accuracy and safety are paramount in Offshore Wind (OSW) maintenance, yet conventional Large Language Models (LLMs) often fail when confronted with highly specialised or unexpected scenarios. We introduce RAGuard, an enhanced Retrieval-Augmented Generation (RAG) framework that explicitly integrates safety-critical documents alongside technical manuals.By issuing parallel queries to two indices and allocating separate retrieval budgets for knowledge and safety, RAGuard guarantees both technical depth and safety coverage. We further develop a SafetyClamp extension that fetches a larger candidate pool, "hard-clamping" exact slot guarantees to safety. We evaluate across sparse (BM25), dense (Dense Passage Retrieval) and hybrid retrieval paradigms, measuring Technical Recall@K and Safety Recall@K. Both proposed extensions of RAG show an increase in Safety Recall@K from almost 0\% in RAG to more than 50\% in RAGuard, while maintaining Technical Recall above 60\%. These results demonstrate that RAGuard and SafetyClamp have the potential to establish a new standard for integrating safety assurance into LLM-powered decision support in critical maintenance contexts.

 \and Retrieval-Augmented Generation (RAG) \and Decision Support \and Safety-critical
\end{abstract}

\section{Introduction}
\label{sec:intro}
Accuracy and safety are paramount in Offshore Wind (OSW) maintenance, an industry characterised by challenging environmental conditions, remote operations, and highly complex technical tasks. Human error or inaccurate decisions during maintenance activities can lead to costly downtime, significant safety risks, and environmental hazards. As global reliance on OSW power grows, ensuring the reliability and safety of these operations becomes critically important.

Recent advances in Artificial Intelligence (AI), particularly Large Language Models (LLMs), offer considerable promise as decision-support tools capable of assisting maintenance personnel by providing immediate access to relevant knowledge. LLMs have demonstrated powerful capabilities to generate context-aware recommendations, summarise complex documents, and support operational decision making. However, conventional LLM-based systems often falter when confronted with unexpected or highly specialised scenarios, primarily due to the limited availability of relevant, scenario-specific training data. This shortcoming is particularly problematic in OSW maintenance, where unexpected scenarios can significantly increase operational risks and decision complexity.

Addressing this challenge, we introduce RAGuard, an enhanced Retrieval-Augmented Generation (RAG) framework explicitly tailored for OSW maintenance contexts. RAGuard integrates in-context safety protocols directly into the retrieval and generation process, dynamically prioritising safety considerations based on real-time maintenance scenarios. Unlike traditional RAG systems, RAGuard leverages specialised technical documentation, such as maintenance manuals and technical data sheets, to retrieve highly relevant, context-specific information precisely when needed. This real-time retrieval process ensures that maintenance guidance remains accurate, current, and scenario-appropriate, significantly reducing the risks associated with unexpected situations.

To systematically assess the effectiveness of RAGuard, we propose an evaluation framework specifically designed to evaluate the safety and reliability of AI-based decision-support systems in OSW maintenance tasks. Preliminary evaluations suggest that RAGuard improves the quality and contextual relevance of maintenance guidance, indicating its potential to enhance operational safety. While further operational validation is required, initial results highlight RAGuard as a promising approach toward safer and more reliable decision-support systems in critical maintenance contexts.

The remainder of this paper is structured as follows: Section \ref{sec:background} reviews relevant literature on RAG methodologies and safety-focused AI. Section \ref{sec:methodology} provides an in-depth description of the RAGuard framework. Section \ref{sec:evaluation} details our experimental methodology and newly proposed benchmark. Section \ref{sec:result} presents our empirical results and their implications, and Section \ref{sec:conclusion} concludes with a discussion of contributions and future research directions.

\section{Background}
\label{sec:background}
\subsection{Retrieval-Augmented Generation}
RAG is a paradigm that integrates information retrieval with generative models to improve knowledge-intensive Natural Language Processing (NLP) tasks. Foundational work by \cite{lewis2020retrieval} introduces RAG models that combine a parametric neural generator with a non-parametric memory (e.g. a Wikipedia index), enabling the generator to retrieve relevant documents and produce more factual, specific answers. This approach achieved state-of-the-art performance on open-domain question answering tasks, outperforming purely parametric models. Recent advancements have further enhanced RAG. For example, \cite{izacard2023atlas} develops Atlas, a pre-trained RAG model that excels in few-shot learning settings. Atlas can attain over 42\% accuracy on the Natural Questions benchmark using only 64 examples, surpassing a 540-billion-parameter closed-book model with far fewer parameters. Such progress demonstrates RAG’s effectiveness in injecting up-to-date knowledge into LLMs while controlling model size and hallucinations.
\subsection{Safety-Focused AI in Critical Domains}
AI systems deployed in safety-critical industries (healthcare, aviation, energy, etc.) must be designed with rigorous safety and reliability considerations. In healthcare, for instance, the use of AI for clinical decision support raises concerns about accountability and patient harm. \cite{habli2020ai} argues that current safety assurance practices are not yet adjusted to AI-driven tools, which can make high-stakes decisions in ways that are opaque to clinicians. They emphasise the need for new frameworks of moral accountability and dynamic safety assurance throughout an AI system’s lifecycle. In aviation, AI and Machine Learning (ML) are being applied to augment safety analysis and risk prediction. A recent systematic review by \cite{demir2024aviation} shows that techniques like deep learning, time-series modelling, and optimisation algorithms are increasingly used to detect patterns in aviation data and enhance safety measures. These AI-driven methods support proactive safety management (e.g. predictive maintenance and improved air traffic control) to help prevent accidents before they occur. In the energy sector, especially nuclear power, AI offers potential to improve monitoring and emergency response in complex industrial systems. \cite{jendoubi2024ai} surveys AI applications in nuclear power plants, noting that AI-based predictive analytics and real-time data processing could bolster reactor safety and decision-making. Their findings highlight early warning systems that use ML and Internet of Things (IoT) sensors to detect anomalies, coordinate with operators, and mitigate risks in critical scenarios. However, they also point out challenges such as the need for updated regulations and cybersecurity safeguards when integrating AI into safety-critical infrastructure.
\subsection{Adaptive Retrieval Mechanisms}
Adaptive retrieval mechanisms dynamically adjust how and when external information is fetched during generation, often making retrieval context-aware or safety-aware. Instead of a fixed one-pass retrieval, these approaches allow the AI to decide if additional knowledge is needed and to retrieve iteratively or on the fly. For example, recent RAG variants like Forward Looking Active REtrieval Augmented Generation (FLARE) and Self-RAG equip language models with the ability to trigger new retrievals based on the model’s internal confidence or reflection tokens. This means the model can autonomously determine the optimal moments to query the knowledge base, stopping when enough information has been gathered. \cite{asai2023self} proposes Self-RAG, where the model “self-reflects” on its draft answer and issues further queries if needed, which streamlines the retrieve-generate loop and improves answer accuracy. Similarly, \cite{jiang2023active} introduces an active retrieval strategy that monitors the generation process and fetches new evidence when the model’s certainty falls below a threshold, thereby tailoring retrieval to the query’s complexity. Beyond research prototypes, the concept of adaptive retrieval is evident in systems like OpenAI’s WebGPT, which used reinforcement learning to train GPT-3 to invoke a search engine mid-generation. This allowed the model to decide when to look up information and even cite sources, behaving like an agent that can use tools. Equally important is making retrieval safety-aware, so that the information brought into the generation loop does not introduce harm or vulnerabilities. One line of work addresses filtering and control of retrieved content. For instance, security evaluations such as SafeRAG \cite{liang2025saferag} show that without safeguards, adversarial or toxic documents can be injected into the retrieval corpus, leading to misleading or harmful outputs. This has underscored the need for robust retrieval filters and context validation. In practice, integrating content moderation\textendash e.g. removing offensive or contradictory results before generation\textendash is becoming a recommended step in RAG pipelines. By dynamically assessing the safety of retrieved passages (using allow-lists, block-lists, or classifier checks), an adaptive retrieval system can reject or down-weight unsafe context. Such context-aware and safety-conscious retrieval mechanisms are an active research area aimed at ensuring that AI systems remain reliable and aligned even as they pull in external information.

In summary, while existing approaches provide robust RAG methods and explore AI safety considerations in critical sectors, there is limited research on unifying these aspects for operational decision support. This paper addresses this gap by presenting RAGuard, a framework that integrates safety-aware retrieval filtering with context-specific generation to enhance reliability in OSW maintenance scenarios.
\section{Methodology}
\label{sec:methodology}
\subsection{Retrieval-Augmented Generation}
RAG enhances LLMs by integrating external knowledge sources at inference time. Unlike traditional LLMs, which rely solely on their internal parameters\textendash often outdated or insufficient for specialised domains, RAG enables access to up-to-date, domain-specific information through retrieval. 

In a typical RAG pipeline, the user's query is embedded into a dense vector using an encoder (e.g., a bi-encoder like Dense Passage Retrieval (DPR)). Meanwhile, the external corpus is pre-encoded into the same vector space and stored in a similarity search index (e.g., Facebook AI Similarity Search (FAISS)). The retriever ranks passages based on similarity (cosine or inner product), and some systems apply additional filtering to discard irrelevant or low-confidence results.

The top-ranked passages are combined with the query to create an enriched prompt. In fusion-in-decoder architectures, passages are encoded separately, and the decoder attends to all inputs to extract relevant information. This retrieval-augmented setup enhances factual grounding and reduces hallucinations while keeping the model compact.

\begin{figure}
\includegraphics[width=0.93\textwidth]{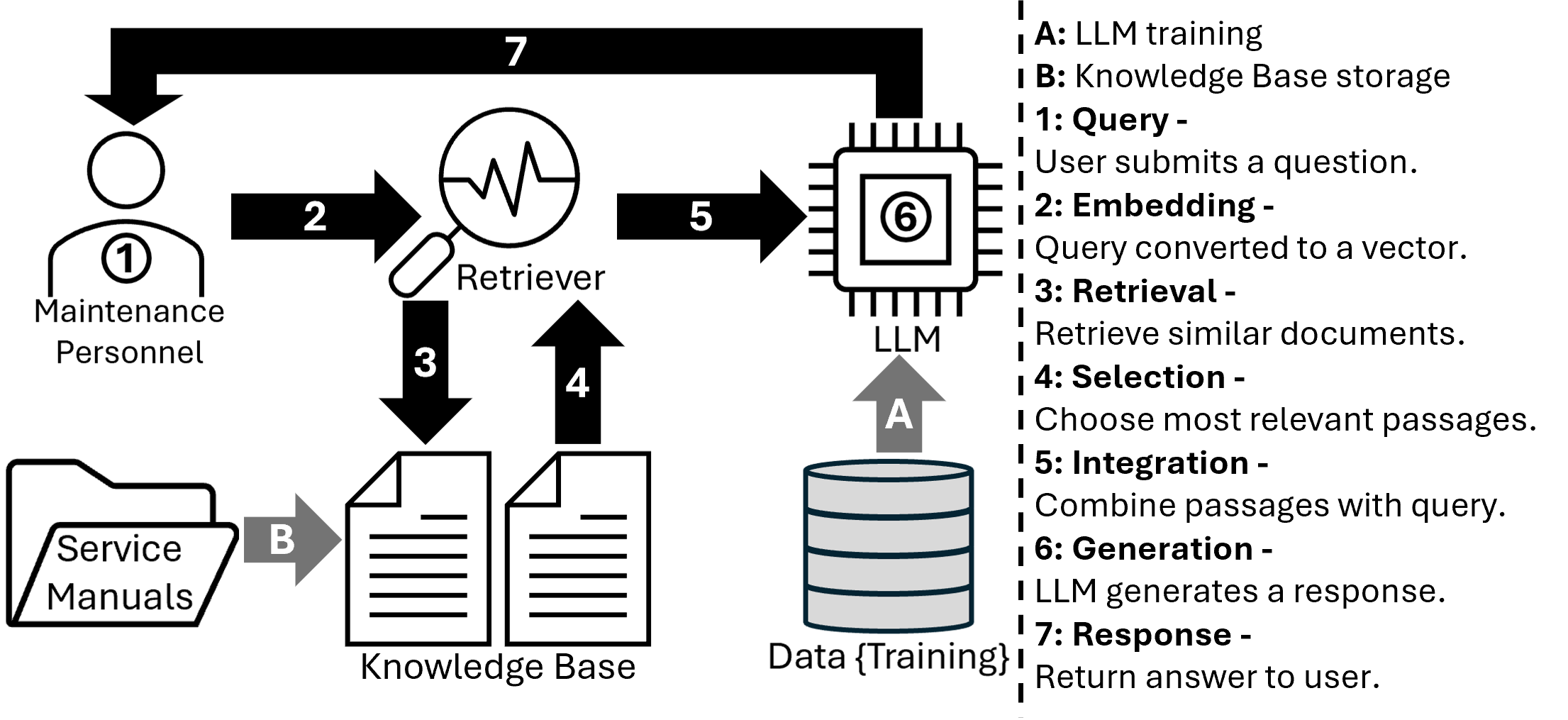}
\caption{Standard RAG Model Process} \label{fig1}
\end{figure}

The generator integrates retrieved content into a coherent, informed response. Incorporating external evidence makes RAG outputs more accurate and context-aware than closed-book approaches. Figure \ref{fig1} summarises each step of the RAG process below.
\subsubsection{RAG Retrieval Parameters}
As with LLMs, RAG systems include tunable hyperparameters that balance retrieval, latency, and noise robustness.

Before indexing, documents are divided into smaller chunks. \textit{Chunk size} controls how much text is embedded at once. Smaller chunks improve retrieval granularity\textemdash helping to match precise content, but increase memory and latency. Larger chunks reduce total passages but may dilute relevance.

\textit{Chunk overlap} determines how much text is shared between adjacent chunks (typically 25-50\% of \textit{chunk size}). It helps preserve context continuity\textemdash preventing important sentences from being split \textemdash at the cost of indexing more passages. Too little overlap risks breaking up meaningful context across chunk boundaries.

\textit{Top-k} specifies how many chunks are returned to the LLM prompt. Higher values increase coverage but raise the chance of irrelevant content and computational load. Lower values reduce noise but may miss critical context. Standard RAG retrieves the $K$ most relevant chunks by maximising the total score:
\begin{equation}
    RAG(q ; K) = \underset{D' \subseteq D,\ |D'| = K}{\arg\max} \underset{d\in D'}{\sum}s(q,d)
\end{equation}
Here, $q$ denotes the query, $D$ is the full document corpus, $D'$ is the selected subset of $K$ passages, and $s(q,d)$ is the similarity score between the query and passage $d$. This optimisation ensures that the top-$K$ passages with the highest aggregated relevance scores are retrieved.

Building on this, we introduce RAGuard which extends the pipeline with explicit safety mechanisms.

\subsection{RAGuard}
Although traditional RAG systems are effective in dynamically integrating external knowledge into generative processes,  they do not explicitly prioritise safety considerations, potentially retrieving contextually relevant but operationally inappropriate or unsafe guidance. To address these limitations, we propose RAGuard, an enhanced RAG framework specifically developed to prioritise safety and contextual accuracy in OSW maintenance operations. The fundamental innovations of RAGuard compared to traditional RAG are safety cache integration, dynamic safety prioritisation, and safety-guided generation. We also propose an additional "SafetyClamp" layer on RAGuard to over-retrieve context passages, reserve predefined knowledge and safety slots, and dynamically fill any leftover slots to supply a full quota.

\subsubsection{Safety Cache Integration}
RAGuard utilises a dedicated cache that contains validated safety protocols and operational guidelines. Unlike a conventional RAG system that draws from a single monolithic document index, RAGuard maintains two parallel knowledge repositories: one containing general maintenance documentation and the other devoted exclusively to safety-critical content. This may include documents such as regulations, industry-specific protocols, and any other relevant information pertinent to the safe completion of all maintenance tasks within the given environment.

At retrieval time, the user's query is issued simultaneously against both corpora, ensuring that the model can draw on rich technical details, while also surfacing explicit hazard warnings, procedural safeguards, and regulatory guidelines. By isolating safety passages in their own index, we can apply dedicated filtering and scoring thresholds that reflect the gravity of risk management, without diluting the coverage or performance of the broader maintenance knowledge base. 

The dual-stream retrieval process produces two ranked lists of passages; one optimised for technical relevance, and the other for safety assurance, which are then merged before context integration. This architecture therefore guarantees that every generated recommendation is grounded not only in accurate domain expertise but also in up-to-date, rigorously validated safety information. We propose two merging functions to this effect and evaluate the effectiveness of both.

\subsubsection{RAGuard Retrieval Parameters}
RAGuard introduces three new retrieval hyperparameters; \textit{knowledge-k} (denoted $k_{know}$), \textit{safety-k} (denoted $k_{safe}$), and \textit{fetch-k} (denoted $k_{fetch}$). These control the balance between technical depth and safety oversight in the dual index setup. The aforementioned $top-k$ remains in use as the total number of passages passed to the prompt context. 

\subsection{RAGuard Retrieval}
RAGuard modifies the standard RAG retrieval step by splitting the total $K$ retrieved passages into two parts: $k_{know}$ from a technical knowledge index, and $k_{safe}$ passages from a safety-specific index. This ensures that the final context includes both technical and safety-relevant content, with $K=k_{know} + k_{safe}$.

At retrieval time, the user's query $q$ is sent to both indices in parallel. Each index returns its top $K$ passages according to a relevance score function $(q, d)$. The two sets of results are then merged into a single list and passed into the LLM prompt. For instance, if $k_{know}=2$ and $k_{safe}=3$, the final prompt includes five passages: two from the knowledge index, and three from the safety index. The process can be formalised as follows: \\
\textbf{Knowledge index retrieval:}
\begin{equation}\label{eq:k_know}
    M_{know}=\{d_1,d_2,\ldots,d_{k_{know}}\},\quad
    d_i = \underset{d\in D_{k_{know}} }{\arg\max}\ s(q,d)
\end{equation}
\textbf{Safety index retrieval:}
\begin{equation}\label{eq:k_safe}
    S_{safe}=\{d'_1,d'_2,\ldots,d'_{k_{safe}}\},\quad
    d'_j=\underset{d\in D_{k_{safe}} }{\arg\max}\ s(q,d)
\end{equation}
\textbf{Combined prompt input:}
\begin{equation}
    RAGuard(q) = [d_1, \ldots, d_{k_{know}},\ d'_1, \ldots, d'_{k_{safe}}]
\end{equation}
In summary, RAGuard queries two indices simultaneously, retrieves fixed top results from each, and merges them into a structured LLM input. The next section introduces \textit{SafetyClamp}, which extends this by enforcing safety quotas and over-retrieving to maximise coverage and control.
\subsection{RAGuard with SafetyClamp}
RAGuard with the additional SafetyClamp builds directly on the base framework by enforcing an absolute safety guarantee on every retrieved passage. Rather than simply interleaving $k_{know}$ and $k_{safe}$ candidates, SafetyClamp begins by over-retrieving a wider pool of contenders; for both the knowledge index and dedicated safety index, the system retrieves the top $k_{fetch}$ passages by relevance. This over-retrieval ensures that, even under strict slot requirements or occasional index sparsity, there will always be enough qualified passages to fill every reserved slot.

Once both pools are retrieved, SafetyClamp assigns passages in a hard-guaranteed sequence. The first $k_{know}$ slots are filled by the highest scoring passages from the knowledge index. Next, the pipeline selects exactly $k_{safe}$ passages from the safety index. Unlike the base RAGuard, $K > k_{know}+k_{safe}$, meaning once this is complete, there are still empty slots for additional passages. These are filled by the combined retrieved pools from both indices, choosing the next highest scoring passages not yet selected, regardless of whether they come from the knowledge or safety index. Given that $k_{fetch}$ exceeds the final $K$, this wildcard mechanism reliably completes the prompt without sacrificing safety guarantees or technical comprehensiveness.

We can formalise RAGuard with SafetyClamp in three steps, reusing $M_{know}$ and $S_{safe}$ defined by equations \ref{eq:k_know} and \ref{eq:k_safe}. First, we over-retrieve a combined candidate list:
\begin{equation}
    C=[c_1,c_2,\ldots,c_{k_{fetch}}]
\end{equation}
Where, $c_i= \underset{d\in D \backslash d\notin \{ c_1,\ldots ,c_{i-1} \} }{\arg\max}\  s(q,d)$\\ \\
Next, we remove any already selected passages from $M_{know} \cup S_{safe}$, preserving order to form the wildcard pool:
\begin{equation}
    R=[r_1,r_2,\ldots]
\end{equation}
Where each $r_l=c_{i_l} \in C$ and $c_{i_l} \notin M_{know} \cup S_{safe}$, and, $i_1 < i_2 < \ldots$ are the indices of those survivors in $C$.\\ \\
Finally, SafetyClamp guarantees exactly $k_{know}$ knowledge passages, $k_{safe}$ safety passages, and fills the remaining ($K-(k_{know}+k_{safe})$) slots from $R$:
\begin{align}
    \text{SafetyClamp}(q; K; &k_{know}; k_{safe}) = \notag \\
    &\quad [m_1, \ldots,m_{k_{know}},s_{k_{safe}}, \ldots,r_1, \ldots,r_{K-(k_{know} + k_{safe})}]
\end{align}
Here $ \{ m_i \} =M_{know}$, $ \{ s'_j \} =S_{safe}$, and the $r$s are drawn from the filtered R. Because $k_{fetch} > K$, there are always enough wildcards to complete the list.  

In essence, SafetyClamp ensures its dual objectives by allocating fixed slots for knowledge and safety passages, over-retrieving extra candidates as wildcards, and assembling the prompt to meet quota requirements and context size. This enforces a safety minimum while preserving technical depth, with over-retrieving preventing empty slots.

\section{Evaluation}
\label{sec:evaluation}
In this section, we describe how we measure each system’s ability to deliver both accurate maintenance guidance and essential safety information under realistic OSW conditions. We first outline our curated evaluation dataset of domain‐specific queries paired with "gold‐standard" answers and regulatory excerpts. We then present the metrics used to quantify technical fidelity, safety compliance, and system efficiency. Finally, we report and analyse results for standard RAG, base RAGuard, and RAGuard with SafetyClamp, highlighting the trade‐offs each design makes between precision, coverage, and latency.
\subsection{Evaluation Dataset}
The evaluation leverages a dataset of 100 maintenance-focused questions, each paired with a "gold-standard" technical answer and the corresponding safety context drawn from two key industry regulations: the Provision and Use of Work Equipment Regulations (PUWER) and the Work at Height Regulations (WAHR). 

For every query, we manually curate the precise procedural steps that constitute the correct technical resolution, and annotated the relevant excerpts from PUWER and WAHR that articulate the required safety checks, hazard warnings, or permitted work practices. This structured format allows us to measure not only whether each system retrieves the passages necessary to reconstruct the technical solution, but also whether it surfaces the exact regulatory language needed to ensure compliance with both PUWER and WAHR. 

By combining domain-specific questions with dual sources of ground-truth (technical and safety), our dataset provides a rigorous test bed for assessing how well RAG, RAGuard and RAGuard with SafetyClamp balance operational accuracy against mandatory safety requirements. 

Crucially, we evaluate each of these three pipelines under all three retrieval paradigms: sparse (BM25), dense (DPR), and hybrid (a weighted fusion of BM25 and DPR scores), to isolate the effect of the underlying retriever on both technical fidelity and safety compliance.
\subsection{LLM Prompt Structure}
Each retrieval pipeline shares a common prompt template that clearly delineates technical guidance from safety considerations. At the top of the prompt, the model is instructed to use the provided context to answer the question and admit when it does not know an answer rather than hallucinating. The template then presents two labelled context sections: under "Maintenance Context", the passages retrieved from the technical knowledge index are inserted, and under "Safety Context", the passages from the safety index appear. As the standard RAG pipeline does not explicitly differentiate between general knowledge and safety, both sections are identical using the $top-k$ passages retrieved from one index. Following these sections, the user's query is posed with a "<<QUESTION>>" marker, ensuring that the LLM's attention remains focused on the specific maintenance task.

Below the question, the prompt ends with a structured "ANSWER" area containing two numbered slots. The first slot, labelled "1) Procedure:," is where the model should generate step-by-step maintenance instructions grounded in the technical context. The second slot, labelled "2) Safety Considerations:," allows explicit hazard checks, warnings, or regulatory requirements drawn from the safety context. 

By splitting the expected output into these two clearly defined components, we can directly assess both the procedural accuracy of the generated guidance and the completeness of the safety advice during evaluation.
\subsection{Evaluation Metrics}
\subsubsection{Hyperparameter Optimisation} 
To establish a fair basis for comparing all pipelines, we first perform a comprehensive hyperparameter selection step. In this, we sweep across the retrieval paradigms and a grid of quota settings ($k_{know}$, $k_{safe}$, $k_{fetch}$, and $K$). 

For each combination, we measure $Retrieval Recall@K$ on both technical and safety passages across 100 questions. This single metric allows identification of which retrieval regime and which hyperparameter values maximise the likelihood of fetching all required "gold-standard" passages. 

Once we determine the best settings for each pipeline, we fix those values for the remaining evaluations. This two-stage approach ensures that all systems are compared under their strongest retrieval settings, yielding a more meaningful assessment of their safety-aware enhancements.
\subsubsection{Retrieval Recall@K}
We measure $Retrieval Recall@K$ separately for technical and safety passages. For each query, we examine the set of passages provided to the LLM and record whether the "gold-standard" technical passage and the annotated safety excerpt appear among the $top-K$. Averaging over all 100 queries yields two recall scores; one reflecting the likelihood of finding the correct procedural context, and one capturing the chance of including at least one requisite safety clause in the prompt.
\subsubsection{Safety Compliance Recall}
This measures whether the retrieved safety passages collectively cover every regulatory requirement specified for a given question. We treat a query as compliant only if all PUWER and WAHR clauses annotated in the dataset appear somewhere in the safety-context feed. The resulting recall rate thus reflects each pipeline's ability to surface the full set of mandated safety checks.
\subsubsection{Latency and Context Utilisation}
To gauge the practicality of the real world, we measure the average end-to-end retrieval time over the full dataset, computing the ratio of occupied to available tokens in the LLM's context window ($K/max-content-size$). This reveals how each approach balances richer contextual grounding against system responsiveness and prompt-size constraints.

We ran all of our latency and context-utilisation measurements on a high-end workstation laptop to approximate a realistic "edge" deployment scenario. The machine is a Ubuntu 22.04.4 LTS system powered by a 13th-Generation Intel\textsuperscript{\textregistered} i9-13980HX (24 threads @ up to 5.6 GHz), with 64 GB of DDR5-5600 RAM and an NVIDIA RTX 4090 GPU. Retrieval timings were recorded on the CPU only; the retriever indices live in memory and are served locally, while the context-window fractions assume a model with a 4,096 token window (4K model). We report both the mean and the standard deviation of 100 runs per pipeline, to display not only the "typical" latency but also its variability under this hardware configuration.
\section{Results \& Discussion}
\label{sec:result}
\subsection{Hyperparameter Optimisation}
We perform a full grid search over our four retrieval hyperparameters—$K$, $k_{know}$, $k_{safe}$, and $k_{fetch}$—subject to $1\leq k_{know}<K$, $1\leq k_{safe}\leq K-k_{know}$, and $K\in \{1,\cdots , 10\}$, $k_{fetch}\in \{25,50,75,100,125,150,175,200\}$, plus the Base RAG cases ($k_{know}=k_{safe}=0, k_{fetch}=\text{None})$.\\If $N$ is the number of distinct $K$ values and $F$ the number of $k_{fetch}$ options, the total number of valid 4-tuples ($K, k_{know}, K_{safe}, k_{fetch}$) we test is:
\begin{equation}
    |\text{settings}|=F\sum_{K=1}^{N}\frac{(K-1)K}{2}+N
\end{equation}
For $N=10$ and $F=8$,this yields 1,330 distinct configurations. For each we compute:
\begin{equation}
    \text{Combined Recall}=\frac{1}{2}(\text{Knowledge Recall + Safety Recall)}
\end{equation}
and then select, for each of the nine pipelines, the configuration that maximises Combined Recall. The resulting optimal parameters and their achieved recall scores are reported in Table \ref{tab:best_combined_recall}.
\begin{table}[ht]
\centering
\begin{threeparttable}
\caption{Best Combined Recall for Each RAG Variant}
\label{tab:best_combined_recall}
\begin{tabular}{l lccccccc}
\toprule
\multirow{2}{*}{\textbf{RAG}} & \multirow{2}{*}{\textbf{Variant}} & 
\multicolumn{4}{c}{\textbf{K Values}} & 
\multicolumn{3}{c}{\textbf{Recall Metrics}} \\
\cmidrule(lr){3-6} \cmidrule(lr){7-9}
& & \textbf{top} & \textbf{know} & \textbf{safe} & \textbf{fetch} & 
\textbf{Knowledge} & \textbf{Safety} & \textbf{Combined} \\
\midrule

\textbf{Dense} & Base    & 10 & -- & -- & --  & 0.925 & 0.09 & 0.508 \\
               & RG\textsuperscript{a}      & 10 & 3  & 7  & --  & 0.535 & 0.92 & 0.728 \\
               & RG-SC\textsuperscript{b}   & 10 & 5  & 5  & 25  & 0.790 & 0.95 & 0.870 \\
\midrule

\textbf{Hybrid} & Base   & 4  & -- & -- & --  & 0.595 & 0.01 & 0.303 \\
                & RG     & 5  & 1  & 4  & --  & 0.380 & 0.74 & 0.560 \\
                & RG-SC  & 7  & 3  & 4  & 25  & 0.585 & 0.71 & 0.648 \\
\midrule

\textbf{Sparse} & Base   & 2  & -- & -- & --  & 0.250 & 0.00 & 0.125 \\
                & RG     & 3  & 1  & 2  & --  & 0.165 & 0.15 & 0.158 \\
                & RG-SC  & 4  & 2  & 2  & 25  & 0.250 & 0.15 & 0.200 \\
\bottomrule
\end{tabular}
\begin{tablenotes}
    \item \textsuperscript{a}RG: RAGuard, \textsuperscript{b}SC: SafetyClamp
\end{tablenotes}
\end{threeparttable}
\end{table}

The hyperparameter sweep clearly illustrates the trade-offs inherent in each pipeline. Base RAG maximises knowledge recall; Dense achieves nearly 93\% correct technical retrieval when $K=10$, but at the cost of almost zero safety coverage. Introducing RAGuard dramatically raises the safety recall to over 90\%, yet reduces the knowledge recall by roughly half, since only three slots are reserved for technical content. RAGuard with SafetyClamp, by contrast, finds a middle ground: by over-retrieving and then guaranteeing both a proportional number of knowledge and safety passages (e.g. $k_{know}=5, k_{safe}=5$ for Dense + RAGuard and SafetyClamp), it retains high safety recall (95\%) while still preserving a strong knowledge recall (79\%), yielding the highest combined score (0.87). 

Hybrid pipelines behave similarly but start from lower base knowledge recall, and sparse pipelines\textendash inherently limited by BM25's coarse matching-cannot exceed ~25\% knowledge retrieval even when safety is ignored. 

Overall, RAGuard with SafetyClamp consistently delivers the best balance, particularly under dense retrieval, by ensuring that neither technical accuracy nor mandated safety context are sacrificed.
\subsection{Retrieval Recall@K}
\begin{figure}[ht]
  \centering
  \includegraphics[width=0.83\textwidth]{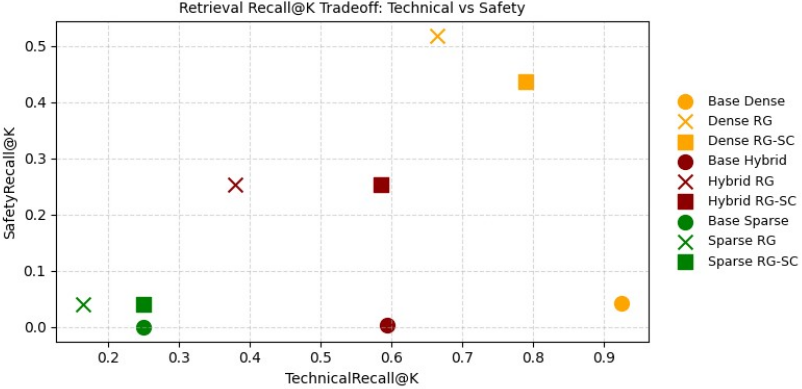}
  \caption{Recall trade-off: Technical Recall@K vs Safety Recall@K}
  \label{fig:recall_scatter}
\end{figure}

Figure \ref{fig:recall_scatter} visualises each pipeline's performance in the technical-vs-safety recall@ K plane, using colour to denote the family (orange: Dense, red: Hybrid, green: Sparse) and distinct markers to indicate the retrieval method ('O': Base RAG, 'X': RAGuard, '$\blacksquare$ ': RAGuard with SafetyClamp). 
\\
The plot reveals a clear three-way trade-off across methods and families. Within each colour band, the Base RAG point sits farthest to the right-maximising technical recall-but remains near the bottom with almost zero safety recall. RAGuard lifts each family sharply upward: for Dense, the jump from Base (0.925, 0.0425) to Dense+RAGuard (0.665, 0.5175) illustrates how interleaving yields large safety gains at the expense of roughly 26 percentage points (pp) of technical recall. SafetyClamp occupies the middle ground, for example, Dense + RAGuard with SafetyClamp at (0.790, 0.4375) recovers most of the Base technical coverage while still boosting safety recall by 39 pp. Hybrid and Sparse variants follow the same pattern; SafetyClamp always improves safety over Base with only a modest drop in technical retrieval, and RAGuard pushes safety even further up. 

Overall, SafetyClamp consistently dominates Base RAG in safety without sacrificing too much technical accuracy, and RAGuard sits at the front of the Pareto curve when safety is paramount. 
\subsection{Safety Compliance Recall}
\begin{figure}[ht]
  \centering
  \includegraphics[width=0.76\textwidth]{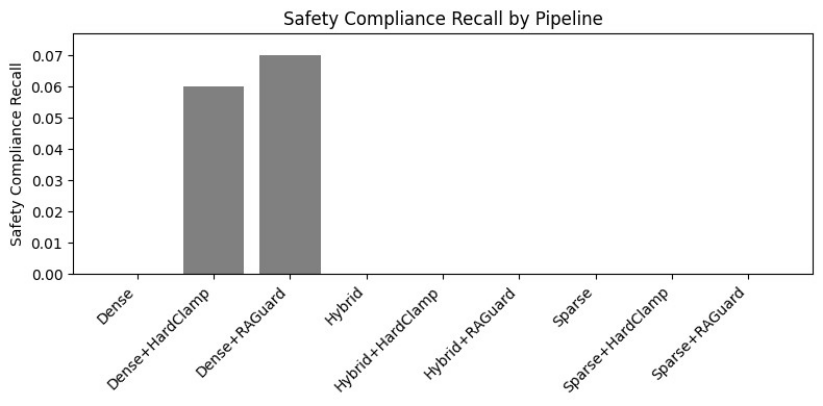}
  \caption{Safety Compliance Recall by Pipeline}
  \label{fig:safety_compliance}
\end{figure}

Figure \ref{fig:safety_compliance} shows each pipeline's Safety Compliance Recall, that is, the fraction of queries for which \textit{all} annotated PUWER and WAHR passages were successfully retrieved. The horizontal axis lists the nine methods, and the vertical axis gives the compliance rate from 0 to 1. Each bar's height corresponds exactly to the Safety Compliance Recall metric: for example, the "Dense + RAGuard" bar at 0.07 indicates that only 7\% of queries retrieved every required safety excerpt under that configuration.
\\
Despite improvements in the single-clause safety recall as presented earlier, full compliance rates remain very low for all pipelines. The best performer, Dense + RAGuard, achieves just 7\% compliance, while its SafetyClamp counterpart achieves 6\%. Hybrid and Sparse variants all fall at or near 0\%, meaning they never retrieve \textit{every} mandated safety clause in a single pass.

These results underscore a critical gap: even the most safety-focused retrieval strategies still omit at least one required regulation clause in over 90\% of cases. To move toward reliable compliance in safety-critical contexts, future work must explore higher safety-slot budgets, multi-pass retrieval until exhaustively covered, or targeted post-retrieval verification steps.
\subsubsection{Latency and Context Utilisation}
\begin{figure}[ht]
  \centering
  \includegraphics[width=0.97\textwidth]{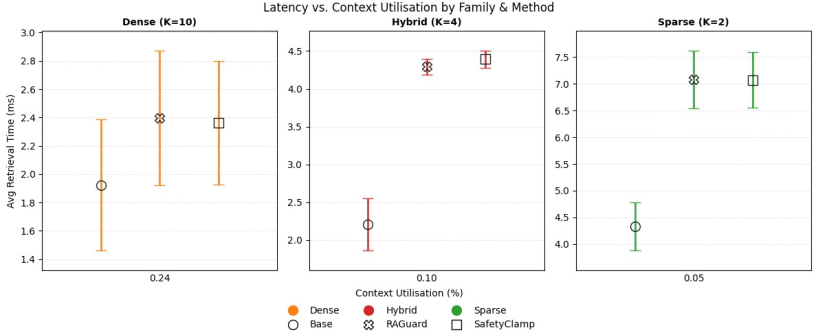}
  \caption{Latency vs Context Utilisation for a 4K Model}
  \label{fig:latency}
\end{figure}

Figure \ref{fig:latency} illustrates the trade-off between retrieval latency and context utilisation across our three RAG families and retrieval methods. Each plot focuses on one family, and plots the average retrieval time on the vertical axis against the fraction of the LLM's context window occupied by retrieved passages on the horizontal axis. Within each plot, a circle denotes the Base RAG, an "X" denotes RAGuard, and a square marks SafetyClamp; all markers are outlined in black, with the error bars in the family's colour showing $\pm$ 1 standard deviation.
\\\\
Across all families, the Base RAG method sits at the leftmost (lowest context use) and lowest latency point. Introducing RAGuard shifts each marker slightly to the right\textemdash because it interleaves safety passages\textemdash at the cost of a modest increase in retrieval time (roughly 0.4-0.6 ms extra). SafetyClamp moves further right, due to its larger $k_{fetch}$, and imposes a further latency penalty; it still completes retrieval in under 3 ms for Dense, under 4.5 ms for Hybrid and under 7.5 ms for Sparse. Thus, the plot makes clear the Pareto front of methods: if the primary goal is minimal latency, and you can tolerate absence of safety guarantees, Base RAG is optimal; if you require safety integration, RAGuard and SafetyClamp offer progressively stronger safety coverage at predictable, bounded increases in retrieval time.
\section{Conclusion}
\label{sec:conclusion}
This work introduced RAGuard, an enhanced RAG framework for OSW maintenance that integrates safety-critical content with technical documentation. By using parallel indices and separate $k_{know}$ and $k_{safe}$ quotas, RAGuard ensures recommendations are grounded in domain expertise and validated safety protocols. We also proposed SafetyClamp, an over-retrieve and hard-clamp variant that guarantees slots for technical and safety passages even when an index is sparse.

Our evaluation, conducted on a curated dataset of 100 real-world OSW maintenance queries, showed that both RAGuard and SafetyClamp substantially outperform standard RAG in surfacing mandated safety clauses. Specifically, Safety Recall@K increased from near 0\% (Base RAG) to over 50\% (Dense RAGuard), with only modest reductions in Technical Recall@K. In hyperparameter sweeps, we found optimal configurations that balance technical fidelity and safety coverage. Latency measurements on a 13th-gen i9 laptop showed that these gains incur only a small retrieval overhead while offering very low context utilisation fractions, leaving ample room in LLM windows.
\sloppy
Overall, RAGuard and its SafetyClamp extension provide a principled, lightweight mechanism for embedding safety guarantees directly into RAG pipelines, offering practical value in regulated high-stakes environments. 

Future work includes several directions. First, we will carry out additional hyperparameter experiments, varying document chunk size and overlap to find the optimal indexing strategy. Second, we plan to integrate more regulatory and technical documents, reflecting the multiple standards and manuals used in real-world operations to ensure our system scales to complex scenarios. We will further investigate adaptive slot-sizing methods that adjust $k_{know}$ and $k_{safe}$ based on the complexity of each query. Finally, we will study how different retrieval configurations influence the quality and safety of the LLM's generated responses. Longer term, we aim to run live field trials to measure the end-to-end effects on maintenance decision accuracy, operational efficiency, and overall safety outcomes, gaining vital expert feedback.

\begin{credits}
\subsubsection{\ackname} This work was conducted under the Aura CDT program, funded by EPSRC and NERC, grant number EP/S023763/1 and project reference 2609857.
\end{credits}
\bibliographystyle{splncs04}
\bibliography{References}
\end{document}